\documentclass[journal]{IEEEtran}
\usepackage[pdftex]{graphicx}
\usepackage{multirow}
\usepackage{makecell}
\usepackage{xcolor}
\usepackage{tabularx}
\usepackage{cite}
\usepackage{amsmath,amssymb,amsfonts}
\usepackage{algorithmic}
\usepackage{graphicx}
\usepackage{textcomp}
\usepackage{xcolor}
\def\BibTeX{{\rm B\kern-.05em{\sc i\kern-.025em b}\kern-.08em
    T\kern-.1667em\lower.7ex\hbox{E}\kern-.125emX}}

\ifCLASSINFOpdf
\else
\fi
\begin{document}
%
\title{
Towards Objective Dysgraphia Detection: A Multi-Branch Deep Learning Approach for Online Handwriting Analysis
}
%
%
%


%
%

\author{\parbox{16cm}{\centering
    {\large Lydia Ouhib$^{1}$, Yassine Ouzar$^{1}$, Zoé Pinseel$^{2}$, 
    Stéphane Bouilland$^{2}$, and Mehdi Ammi$^{1}$}\\
    {\normalsize
    $^1$ LIASD Laboratory, University of Paris 8, Saint-Denis, France\\
    $^2$Centre Jacques Calvé, Fondation Hopale, Berck, France
}
}}

\IEEEpubid{\makebox[\columnwidth]{979-8-3195-1142-3/26/\$31.00 ©2026 IEEE \hfill}%
\hspace{\columnsep}\makebox[\columnwidth]{}}

\maketitle 

\IEEEpubidadjcol


%
\IEEEpeerreviewmaketitle

\begin{abstract}

Dysgraphia is a specific learning disability that is prevalent among school-age children. It affects handwriting coherence, quality, fluency, and legibility, often hindering academic achievement and early learning development. This motor coordination disorder is typically diagnosed through subjective assessments based on clinician observation, which can be time-consuming and prone to variability. 
In this paper, we introduce a deep learning-based framework for objective dysgraphia detection using online handwriting data captured via digitizing tablets. The proposed framework relies on two complementary branches: the first pipeline extracts both handcrafted and embedding-based kinematic features directly from raw temporal signals, while the second leverages image-based representations of the temporal signals generated using continuous wavelet transforms (CWT) and Gramian Angular Fields (GAF). The resulting features are then fused to leverage the complementary strengths of both representations. The four representations were evaluated separately and jointly using the publicly available DiaGraMo dataset, showing that the fusion of GAF, MOMENT, and hand-crafted kinematic features outperforms each individual representation, as well as other fusion schemes. These findings highlight the potential of the complementarity of image and signal based representations for more objective dysgraphia detection.

\end{abstract}


\section{Introduction}
%
%
%
%

Motor coordination refers to the set of processes through which the nervous system integrates signals from the brain, spinal cord, and peripheral nerves to control the timing, force, and precision of muscle activity, enabling smooth and accurate movements. This integration involves continuous feedback and feedforward communication between sensory systems and motor pathways, allowing even the simplest movements to be executed efficiently and adaptively \cite{ting2007neuromechanics}.

Motor coordination disorder affects about 5–6\% of school-aged children. Despite the absence of any apparent neurological lesion, it nevertheless leads to significant and persistent difficulties in acquiring motor skills \cite{pergantis2026artificial}. The development of motor coordination in children is essential for performing daily activities such as writing, drawing, or manipulating objects. Despite the prevalence of this disorder and its long-term impact on quality of life, it remains underdiagnosed in many cases. Current clinical assessment relies on standardized motor test batteries requiring qualified specialists, and is costly, time-consuming, and inaccessible in many educational settings \cite{pergantis2026artificial}. 

Motor coordination disorders may also contribute to learning difficulties affecting writing, such as dysgraphia, which is a specific learning disorder characterized by slow, effortful, and illegible script \cite{mekyska2016identification, gargot2020acquisition}. Handwriting analysis has emerged as a promising biomarker for early detection of motor and neuro-developmental disorders \cite{mekyska2016identification}. Children with developmental coordination disorder and attention deficit hyperactive disorders show distinctive kinematic signatures in their handwriting : irregular velocity profiles, unpredictable acceleration peaks, excessive pen lifts, and unstable pressure patterns all of which can be objectively captured by digitizing tablets \cite{mekyska2016identification, asselborn2018automated}. 

Current dysgraphia diagnosis is mainly based on clinical observation which is subjective, time-consuming, and varies between clinicians. Faced with these limitations, several studies have sought to automate dysgraphia assessment by using scanned paper sheets of handwriting \cite{weraduwa2024developing, ramlan2023potential}. While these approaches are promising and have demonstrated their ability to identify certain visual markers of dysgraphia, they rely solely on the final image of the writing. Consequently, they neglect temporal and kinematic information related to the child's movements during writing. Several recent studies have also proposed classical machine learning models based on hand-crafted feature extraction, often applied to private databases \cite{drotar2020dysgraphia, gargot2020acquisition}. However, these approaches are heavily dependent on feature engineering, thus limiting their generalizability and adaptability to diverse contexts.

To address these drawbacks, we developed a multi-branch deep learning model for dysgraphia detection from online handwriting Data. The first branch extracts both handcrafted and embedding-based kinematic features directly from raw handwriting temporal signals, while the second leverages image-based representations of the temporal signals generated using continuous wavelet transforms (CWT) and Gramian Angular Fields (GAF). The resulting features are then fused to leverage the complementary strengths of both representations. Both pipelines were evaluated separately and jointly on the new publicly available DiaGraMo dataset.


The remainder of this paper is organized as follows. Section II presents the materials and methods. Section III reports and discusses the results. Finally, Section IV concludes the paper and highlights future research directions.

\begin{figure*}[h]
\begin{center}
\includegraphics[scale=0.4]{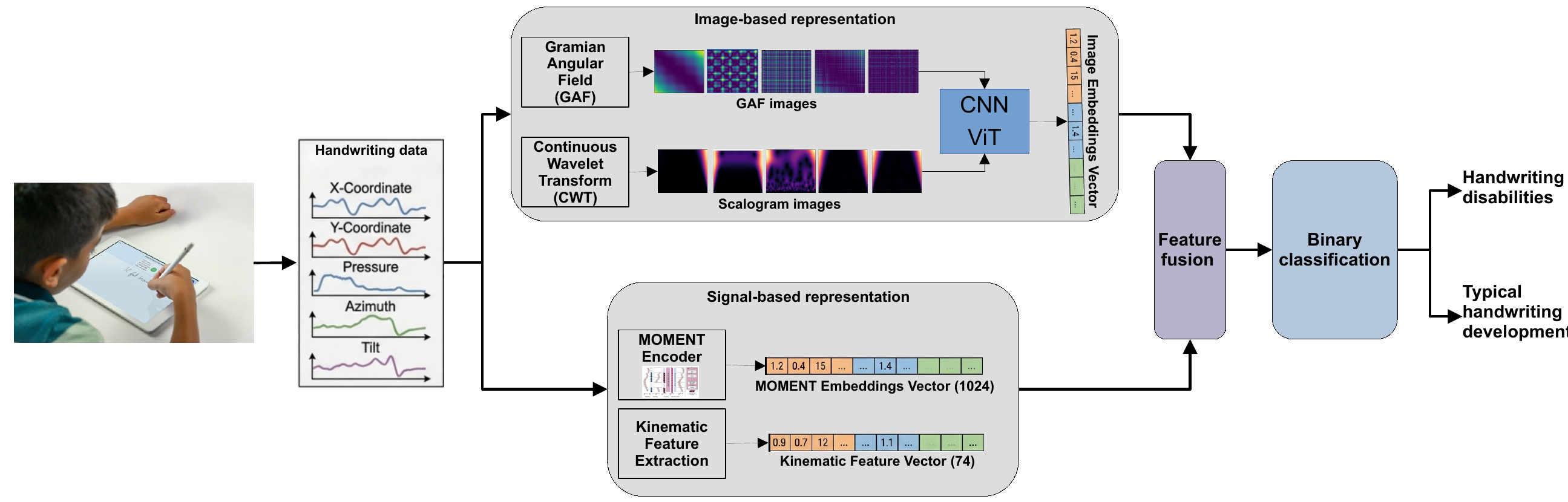}

\end{center}
   \caption{Overview of the proposed multimodal framework for dysgraphia detection from online handwriting signals. 
}
\label{framework}
\end{figure*}
\section{Materials and Methods}
\subsection{Dataset}\label{AA}

Although several handwriting datasets have been proposed for the study of motor disorders in children, most of them are limited in terms of the type of data they provide. Existing public datasets such as the Sri Lankan numeric dysgraphia dataset \cite{silvadeveloping} and the Malaysian potential dysgraphia dataset \cite{ramlan2023potential} consist exclusively of scanned paper images collected without any recording device. Consequently, they lack essential kinematic data, making them unsuitable for motor behavior analysis.

To the best of our knowledge, Multimodal Czech Online Handwriting and Cognitive Data from Children with and without Handwriting Disabilities referred to as DiaGraMo \cite{zvonvcakova2026multimodal} is currently the only publicly available dataset that provides raw kinematic data derived from children’s online handwriting signals. DiaGraMo consists of kinematic handwriting data captured on digital tablets along with standardized assessments of children's cognitive, visuospatial, phonological, and writing abilities. In this article, we only used the kinematic data of the handwriting that was recorded at 167 Hz using a Wacom Cintiq 16 tablet with a Wacom Pro Pen 2. Overall, the DiaGraMo dataset comprises writing data from 276 young Czech children aged 8 to 12, including 161 child diagnosed with dysgraphia. Each child performed 14 writing tasks, including spirals, zigzags, and loops, a dictation and a text copy, producing seven signals (x coordinate, y coordinate, pressure, tilt, azimuth, timestamp, and pen status).


\subsection{Data Preparation}

The DiaGraMo dataset provides recordings in several formats, including integrated and preprocessed JSON files available in the clean\_data folder. We directly used these files, which were already cleaned and verified by the dataset authors, including the removal of corrupted recordings. Each participant has a merged file containing the seven signals recorded by the tablet.
In order to ensure a balanced dataset at the task level, we selected only the participants who completed all 14 tasks, resulting in a total of 258 participants.

The 14 graphomotor tasks cover a wide range of writing activities, such as fast and precise spiral drawing, loop writing, rainbow patterns, saw-tooth strokes, dictation, and text copying. All available tasks per participant were retained and processed simultaneously, allowing the model to capture the complete handwriting profile of each participant.


\subsection{Overall Framework}

The proposed framework for multimodal dysgraphia detection from online handwriting signals is illustrated in Figure \ref{framework}. We treat this task as a binary classification problem. The model processes the raw kinematic handwriting signals acquired via a digital tablet and learns complementary representations through two parallel branches. The first branch generates image-based representations using Continuous Wavelet Transform (CWT) and Gramian Angular Field (GAF), from which visual features are extracted. The second branch constructs signal-level representations using either learned features from a pre-trained time-series foundation model or domain-specific hand-crafted kinematic features. The resulting features from both branches are then fused to form a unified representation, which is finally used for dysgraphia detection.

\subsubsection{Signal-based Representations}

The first branch operates directly on raw handwriting time series by extracting two signal-based representations using complementary approaches. 

Following the work of Drotar et al. \cite{drotar2020dysgraphia}, the first pipeline extracts 74 hand-crafted kinematic features from the seven recorded handwriting signals, capturing relevant characteristics of handwriting movements associated with motor coordination disorders. These features are grouped into eleven categories: velocity (8 features), acceleration (6 features), jerk (3 features), velocity peaks (3 features), pen lifts (3 features), pressure (8 features), pen orientation (4 features), spatial and temporal trajectory (8 features), curvature (2 features), statistical descriptors including entropy, skewness and kurtosis (10 features), and finally frequency and energy descriptors based on FFT (Fast Fourier Transform) and RMS (Root Mean Square)  (19 features). Overall, the 74 features are extracted for each task and concatenated across 14 tasks per participant, resulting in a 1036 dimensional feature vector per child. This enables the extraction of discriminative representations of writing patterns across tasks.

The second pipeline relies on MOMENT, a pre-trained time series foundation model \cite{goswami2024moment}. Each handwriting signal is resampled to 512 time steps using one-dimensional linear interpolation to fit with the model's fixed input requirement. MOMENT encodes each task into a 1024 dimensional latent representation through its Transformer encoder. For each participant, the latent representations from the 14 tasks are concatenated to form a unified feature vector used for classification.



\subsubsection{Image-based Representations}

Handwriting time series typically exhibit high variability and non-stationary dynamics over time, making direct processing of raw signals challenging. To address this, we also adopt image-based representations of the kinematic signals, which offer two key advantages. First, they encode 
temporal dynamics into a structured spatial form, making the signal more robust to noise and variability. 
Second, they enable the use of powerful pretrained vision encoders such as CNNs and Vision Transformers (ViTs), which have demonstrated state-of-the-art performance in extracting rich and discriminative features from images. Two complementary transformations were selected for this purpose: the Gramian Angular Field (GAF) and the Continuous Wavelet Transform (CWT).



\paragraph{Gramian Angular Field (GAF)}

GAF transforms each time series into a matrix of $224 \times 224$  where each element is the cosine of the sum of angles between two time points : 
\begin{equation}
G_{i,j} = \cos(\phi_i + \phi_j)
\end{equation}
which allows the encoding of global temporal dependencies and recurrence patterns within the signal. A regular and fluid signal, as produced by a typically developing child, will produce a smooth and structured GAF image, reflecting consistent movement patterns. In contrast, an irregular signal with tremors and hesitations, typical of dysgraphic children, will produce a less structured image that reflects the lack of stable temporal patterns.



\medskip

\paragraph{Continuous Wavelet Transform (CWT)}
The CWT provides a time–frequency representation of the signal, allowing the analysis of both slow and fast variations over time. Unlike GAF, which captures global structure, CWT focuses on local changes in the signal. This makes it particularly useful for detecting fine motor irregularities such as tremors, sudden changes in pressure, or variations in writing speed that are often observed in dysgraphia. High-frequency components correspond to rapid and localized variations, while low-frequency components reflect slower and more global movements. In this work, the CWT is computed using a Morlet wavelet for each signal, resulting in a $224 \times 224$ matrix representation.



For each participant and task, the seven $224 \times 224$ matrices produced by GAF or Wavelet transformations are stacked into a single $7 \times 224 \times 224$ tensor, which is then fed into a pretrained deep learning encoder. Four state-of-the-art encoders are evaluated, including ResNet50, EfficientNet-B4, ViT, and VGG16, each of which extracts a compact feature representation from the input tensor. The resulting embeddings are concatenated across all tasks to form a global participant-level representation, which is subsequently used for dysgraphia classification.



\subsection{Multimodal Fusion}

Combining multimodal representations has been shown to improve classification performance in many biomedical tasks, as each representation captures different and complementary characteristics. In our case, GAF and Wavelet encode the temporal dynamics of the handwriting signals into structured visual representations, capturing global correlations and local time-frequency variations respectively, while kinematic features describe the dynamics of handwriting, including movement speed, pressure, and pen-lift frequency. These are key indicators used by clinicians for evaluating handwriting quality and diagnosing dysgraphia.



Building on these considerations, we propose a feature-level multimodal fusion approach in which signal- and image-based representations are concatenated into a unified feature vector. The same set of classifiers is trained independently on each fused configuration to ensure fair and consistent evaluation. We consider eleven fusion settings in total: six pairwise, four three-modality, and one four-modality combination. 







\section{Results and Discussion}



In this section, we present and discuss the experimental results obtained from the different pipelines evaluated in our framework. We conduct a comprehensive analysis of both individual and fused modalities, considering hand-crafted kinematic features, latent kinematic representations encoded by MOMENT, and image-based representations derived from GAF and CWT transformations. All experiments are conducted on the DiaGraMo dataset using a six-fold subject-independent cross-validation strategy, which prevents any data leakage between training and testing sets. Performance is assessed using accuracy, F1-score, and AUC-ROC.
For each representation, the three best-performing encoder–classifier combinations, based on accuracy, are reported in the corresponding tables.

\begin{table}[ht]
\centering
\caption{Results for Signal-based Representations on DiaGraMo.}
\label{tab:raw_signal}
\renewcommand{\arraystretch}{1.2}
\setlength{\tabcolsep}{12pt}
\begin{tabular}{lllccc}
\hline
\textbf{Features} & \textbf{Classifier} & \textbf{Acc} & \textbf{F1} & \textbf{AUC} \\
\hline
\multirow{6}{*}{\raisebox{8ex}{MOMENT}}
 & Logistic Reg        & 65.1 & 65.1 & 70.1 \\ & Random Forest                  & 65.1 & 63.3 & 71.4 \\
 & \textbf{SVM}   & \textbf{70.2}  & \textbf{69.9} & \textbf{74.9} \\

\hline
\multirow{7}{*}{\raisebox{11ex}{Kinematic}} & XGBoost        & 82.2 & 81.9 & 89.9 \\ & LightGBM       & 82.9 & 82.8 & 90.5 \\ & \textbf{AdaBoost}         & \textbf{83.7} & \textbf{83.6} & \textbf{90.1} \\

\hline
\end{tabular}

\end{table}


We first evaluate each representation independently to assess its discriminative capability using several traditional machine learning classifiers, including SVM, Random Forest, XGBoost, LightGBM, AdaBoost, Logistic Regression and MLP. 

Tables ~\ref{tab:raw_signal} and ~\ref{tab:image_based} report the results of unimodal representations, where hand-crafted kinematic features consistently outperform all other unimodal approaches, achieving up to 83.7\% accuracy and 90.1\% AUC with AdaBoost. This result highlights the effectiveness of kinematic features in capturing meaningful motor characteristics of writing activity. However, MOMENT-based representations show lower performance, where SVM reaches 70.2\% accuracy and 74.9\% AUC. This may be due to the fact that the MOMENT foundation model is pre-trained on general time-series data and is not specifically adapted to handwriting data.



\begin{table}[hbtp]
\centering
\caption{Results for Image-Based Representations on DiaGraMo.}
\label{tab:image_based}

\renewcommand{\arraystretch}{1.2}

\begin{tabularx}{\columnwidth}{llXXX}
\hline
\textbf{Encoder} & \textbf{Classifier} & \textbf{Acc} & \textbf{F1} & \textbf{AUC} \\
\hline

\multicolumn{5}{l}{\textit{\textbf{Gramian Angular Field (GAF)}}} \\
\hline
EfficientNet  & XGBoost        & 65.9 & 65.3 & 71.8 \\
VGG16         & Random Forest  & 68.6 & 67.0 & 72.2 \\
\textbf{ViT}           & \textbf{LightGBM}       & \textbf{73.3} & \textbf{73.0} & \textbf{78.5} \\
\hline

\multicolumn{5}{l}{\textit{\textbf{Wavelet Transform (CWT)}}} \\
\hline
VGG16         & SVM            & 66.3 & 66.0 & 70.8 \\
ViT           & SVM            & 67.1 & 66.4 & 72.1 \\
\textbf{EfficientNet}  & \textbf{LightGBM}       & \textbf{71.7} & \textbf{71.5} & \textbf{75.3} \\
\hline

\end{tabularx}
\end{table}

Regarding image-based representations, the combination of GAF with a ViT encoder and LightGBM classifier yields the best performance among all evaluated configurations, reaching an accuracy of 73.3\% and an AUC of 78.5\%. In comparison, the Wavelet representation combined with EfficientNet and LightGBM achieves an accuracy of 71.7\% and an AUC of 75.3\%, which remains below the best GAF-based result. This suggests that encoding global temporal correlations is more discriminative than time-frequency decomposition for dysgraphia detection. However, both image-based representations underperform compared to the kinematic baseline, suggesting that using visual representations alone is insufficient to fully capture the motor characteristics of dysgraphia.



We further evaluate eleven feature-level fusion configurations integrating handcrafted kinematic features, MOMENT embeddings, and image-based representations derived from GAF and CWT transformations. Table~\ref{tab:multimodal} reports the three best-performing multimodal fusion configurations out of the eleven combinations evaluated, covering two-modality, three-modality, and four-modality combinations. The combination of Kinematic + GAF achieves an accuracy of 84.9\% and an AUC of 91.1\%, outperforming the best unimodal performance obtained using kinematic features alone. This improvement confirms that GAF representations provide complementary information to kinematic features by capturing global temporal dependencies in writing dynamics that are not explicitly encoded in handcrafted kinematic features.

Integrating the MOMENT representation into the Kinematic + GAF combination further improves accuracy to 85.3\%, yielding the best overall performance across all experiments. This highlights that the three modalities capture different and complementary aspects of the handwriting. Kinematic features describe directly how the writing movement is performed over time, including velocity, pressure, and pen lift frequency. GAF encodes the global temporal structure of each signal and highlights the regularity or irregularity of handwriting movements, which are key indicators of motor difficulties in dysgraphic handwriting. Furthermore, MOMENT automatically extracts learned embeddings from the raw handwriting signals, providing a complementary view that is not captured by the other representations. This highlights the importance of integrating both signal-based and image-based representations to achieve more robust and discriminative models.


\begin{table}
\centering
\caption{Results for Multimodal Fusion on DiaGraMo.}
\label{tab:multimodal}
\renewcommand{\arraystretch}{1.2}
\resizebox{\columnwidth}{!}{%
\begin{tabular}{llccc}
\hline
\textbf{Modalities} & \textbf{Classifier} & \textbf{Acc} & \textbf{F1} & \textbf{AUC} \\
\hline
Kinematic + GAF & \multirow{2}{*}{AdaBoost} & \multirow{2}{*}{83.7} & \multirow{2}{*}{83.7} & \multirow{2}{*}{90.1} \\
+ Wavelet + MOMENT & & & & \\
\cline{1-5}
GAF + Kinematic & AdaBoost & 84.9 & 84.9 & 91.1 \\
\cline{1-5}
\textbf{Kinematic + GAF + MOMENT} & \textbf{AdaBoost} & \textbf{85.3} & \textbf{85.2} & \textbf{90.8} \\
\hline
\end{tabular}%
}
\end{table}



\section{Conclusions and future works}

In this paper, we introduced a multi-branch deep learning framework for objective dysgraphia detection from online handwriting data. The proposed approach leverages both signal-based and image-based representations to exploit their complementary strengths. Experiments on the publicly available DiaGraMo dataset demonstrate that the fusion of GAF, MOMENT, and hand-crafted kinematic features outperforms both individual representations and other fusion schemes.

Several directions for future work can be considered. First, exploring and fine-tuning pre-trained models that are better suited to handwriting motor data could further enhance the effectiveness of the extracted features. Second, we aim to investigate feature selection strategies based on attention mechanisms, along with more advanced fusion techniques, to better exploit the complementarity of multimodal representation. Finally, the DiaGraMo dataset also includes cognitive test scores for each participant. Exploiting these data as additional features could provide a better understanding of the child’s motor and cognitive profile and improve both the accuracy and interpretability of dysgraphia detection.



\bibliographystyle{unsrt} 
\bibliography{biblio}

\begin{thebibliography}{10}

\bibitem{ting2007neuromechanics}
Lena~H Ting and J~Lucas McKay.
\newblock Neuromechanics of muscle synergies for posture and movement.
\newblock {\em Current opinion in neurobiology}, 17(6):622--628, 2007.

\bibitem{pergantis2026artificial}
Pantelis Pergantis, Konstantinos Georgiou, Nikolaos Bardis, Charalabos Skianis, and Athanasios Drigas.
\newblock Artificial intelligence in the evaluation and intervention of developmental coordination disorder: A scoping review of methods, clinical purposes, and future directions.
\newblock {\em Children}, 13(2):161, 2026.

\bibitem{mekyska2016identification}
Jiri Mekyska, Marcos Faundez-Zanuy, Zdenek Mzourek, Zoltan Galaz, Zdenek Smekal, and Sara Rosenblum.
\newblock Identification and rating of developmental dysgraphia by handwriting analysis.
\newblock {\em IEEE Transactions on Human-Machine Systems}, 47(2):235--248, 2016.

\bibitem{gargot2020acquisition}
Thomas Gargot, Thibault Asselborn, Hugues Pellerin, Ingrid Zammouri, Salvatore M.~Anzalone, Laurence Casteran, Wafa Johal, Pierre Dillenbourg, David Cohen, and Caroline Jolly.
\newblock Acquisition of handwriting in children with and without dysgraphia: A computational approach.
\newblock {\em PloS one}, 15(9):e0237575, 2020.

\bibitem{asselborn2018automated}
Thibault Asselborn, Thomas Gargot, {\L}ukasz Kidzi{\'n}ski, Wafa Johal, David Cohen, Caroline Jolly, and Pierre Dillenbourg.
\newblock Automated human-level diagnosis of dysgraphia using a consumer tablet.
\newblock {\em NPJ digital medicine}, 1(1):42, 2018.

\bibitem{weraduwa2024developing}
Sandushi Weraduwa, Dinesh Asanka, Thilini Mahanama, and Swarna Wijeatunge.
\newblock Developing a dysgraphia handwriting dataset for early detection of dysgraphia in sinhala-speaking children.
\newblock In {\em 2024 9th International Conference on Information Technology Research (ICITR)}, pages 1--6. IEEE, 2024.

\bibitem{ramlan2023potential}
Siti~Azura Ramlan.
\newblock Potential dysgraphia handwriting dataset of school-age children.
\newblock 2023.

\bibitem{drotar2020dysgraphia}
Peter Drot{\'a}r and Marek Dobe{\v{s}}.
\newblock Dysgraphia detection through machine learning.
\newblock {\em Scientific reports}, 10(1):21541, 2020.

\bibitem{silvadeveloping}
GVSC Silva, AGCS Bandara, MWP Maduranga, and M~Imran Uvais.
\newblock Developing a comprehensive handwriting dataset for early detection of numeric dysgraphia across sri lanka.

\bibitem{zvonvcakova2026multimodal}
Katar{\'\i}na Zvon{\v{c}}{\'a}kov{\'a}, Jiri Mekyska, Adam Klocek, Jan Mucha, and Zoltan Galaz.
\newblock Multimodal czech online handwriting and cognitive data from children with and without handwriting disabilities.
\newblock 2026.

\bibitem{goswami2024moment}
Mononito Goswami, Konrad Szafer, Arjun Choudhry, Yifu Cai, Shuo Li, and Artur Dubrawski.
\newblock Moment: A family of open time-series foundation models.
\newblock In {\em International Conference on Machine Learning}, 2024.

\end{thebibliography}

\end{document}